\documentclass[pdflatex,sn-mathphys-num]{sn-jnl}% Math and Physical Sciences Numbered Reference Style
\usepackage{graphicx}%
\usepackage{multirow}%
\usepackage{amsmath,amssymb,amsfonts}%
\usepackage{amsthm}%
\usepackage{mathrsfs}%
\usepackage[title]{appendix}%
\usepackage{xcolor}%
\usepackage{textcomp}%
\usepackage{manyfoot}%
\usepackage{booktabs}%
\usepackage{algorithm}%
\usepackage{algorithmicx}%
\usepackage{algpseudocode}%
\usepackage{listings}%
\usepackage{subcaption}%
\usepackage{bm}%
\theoremstyle{thmstyleone}%
\theoremstyle{thmstyletwo}%

\theoremstyle{thmstylethree}%

\begin{document}

\title[Article Title]{From Answers to Rationales: Self-Aligning Multimodal Reasoning with Answer-Oriented Chain-of-Thought}


\author[1,2]{\fnm{Wentao} \sur{Tan}}\email{ftwentaotan@mail.scut.edu.cn}

\author*[2]{\fnm{Qiong} \sur{Cao}}\email{caoqiong1@jd.com}
\equalcont{Project Lead.}

\author*[2]{\fnm{Yibing} \sur{Zhan}}\email{zybjy@mail.ustc.edu.cn}

\author[2]{\fnm{Chao} \sur{Xue}}\email{xuechao19@jd.com}

\author[1,3]{\fnm{Changxing} \sur{Ding}}\email{chxding@scut.edu.cn}

\affil[1]{\orgname{South China University of Technology}, \orgaddress{\street{381 Wushan Road, Tianhe District}, \city{GuangZhou}, \postcode{510000}, \state{GuangDong}, \country{China}}}

\affil[2]{\orgdiv{JD Explore Academy}, \orgname{JD.com}, \orgaddress{\street{Kexin 11th Street, Yizhuang Economic and Technological Development Zone}, \postcode{100176}, \state{BeiJing}, \country{China}}}

\affil[3]{\orgdiv{Pazhou Lab}, \orgaddress{\street{Artificial Intelligence and Digital Economy Pilot Zone Core Area}, \city{Guangzhou}, \postcode{510335}, \state{GuangDong}, \country{China}}}


\begin{figure*}[b]

\centerline{\includegraphics[width=1.0\linewidth]{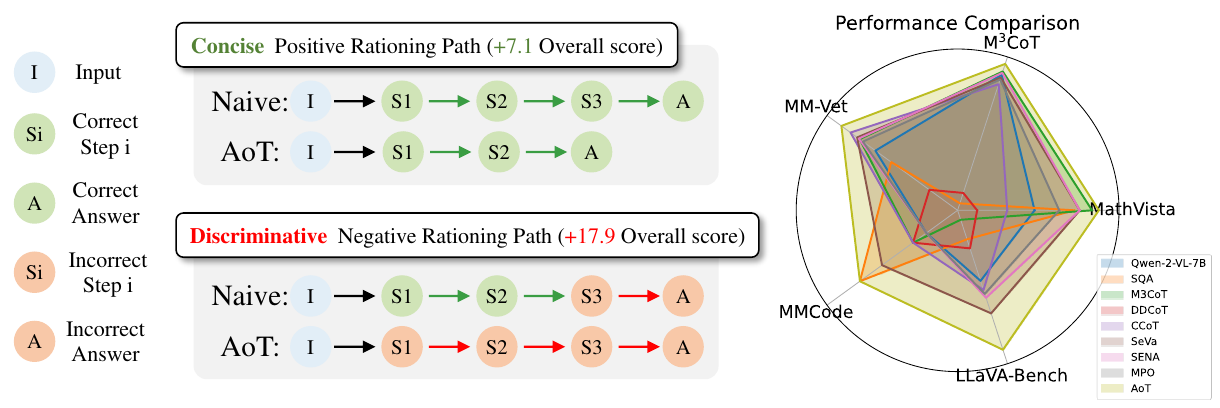}}

  \caption{The left figure summarizes AoT-generated data characteristics, featuring concise positive reasoning and highly discriminative negative reasoning, while the right figure displays the performance of the Qwen2-VL-7B model trained using various automated data generation techniques. AoT delivers the best results.
}
  \label{fig:ijcv}

\end{figure*}

\abstract{Achieving human-like reasoning capabilities in Multimodal Large Language Models (MLLMs) has long been a goal. Current methods primarily focus on synthesizing positive rationales, typically relying on manual annotations or complex systems. Moreover, they often overlook negative reasoning, which limits the model's generalization ability and robustness in multimodal inference.
To address this gap, we propose a novel framework: \textbf{S}elf-Aligning \textbf{M}ultimodal Reasoning with \textbf{A}nswer-O\textbf{r}iented Chain-of-\textbf{T}hought (SMART). SMART employs an answer-oriented chain-of-thought (AoT) prompt to automatically construct high-quality data. Drawing inspiration from human proof-based strategies, AoT leverages both correct and incorrect answers to extract key visual information that links questions and answers. When provided with correct answers, the model produces strong positive rationales. Conversely, when correct answers are replaced with incorrect alternatives, the model generates an erroneous yet compelling reasoning path, serving as a form of discriminative negative rationale. Models trained with AoT-generated data outperform those trained on manually annotated datasets, demonstrating superior reasoning capabilities.
Consequently, SMART establishes an iterative generation-optimization method that continually enhances the model's reasoning skills. Experiments indicate that the SMART framework significantly improves various MLLMs, regardless of model architecture, parameter size, or pre-training dataset.  
The code is available at https://github.com/WentaoTan/SMART. }




\maketitle

\section{Introduction}
\label{sec:intro}

Recently, there has been significant progress in Multimodal Large Language Models (MLLMs) \cite{chen2023minigpt,chen2023shikra,peng2023kosmos,hong2024cogvlm2,li2023blip,chen2024internvl,ye2024mplug,sphinx}. Many impressive visual-text MLLMs \cite{chen2024far,lu2024deepseek,young2024yi,liu2024llavanext,yao2024minicpm,Qwen2VL} have emerged, demonstrating excellent performance in tasks like image captioning \cite{wang2023llm,liu2023llava} and visual question answering \cite{fu2023mme,antol2015vqa,li2024seed}. However, as task complexity increases, these models reveal limitations in their reasoning abilities. For example, while they perform well on simple benchmarks, they struggle with more complex tasks that require logical reasoning \cite{yu2023mm,lu2024mathvista,chen2024m}. Developing AI systems capable of complex multimodal reasoning, akin to human cognition, is a key objective in the MLLM field. Therefore, enhancing the reasoning capabilities of these models is of utmost importance.

\begin{figure*}[t]

\centerline{\includegraphics[width=1.0\linewidth]{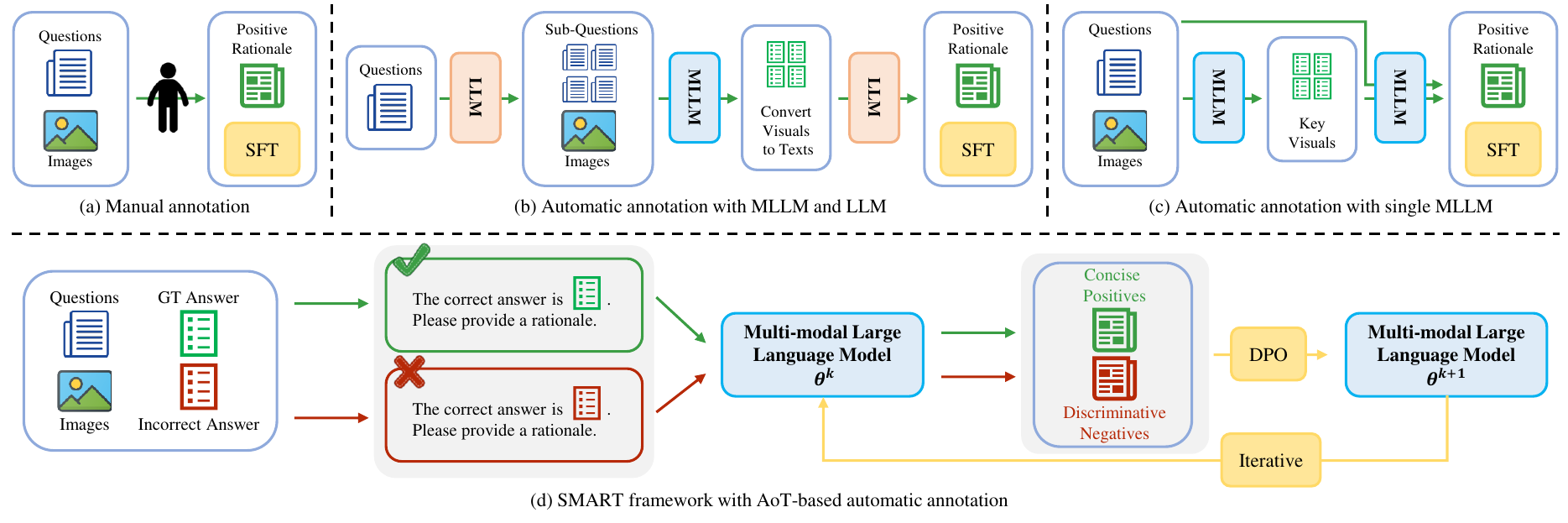}}

  \caption{
  An overview of existing annotation methods: (a) manual annotations for creating positive rationales \cite{lu2022learn,chen2024m}, (b) a combined LLM and MLLM approach for generating reasoning datasets \cite{zheng2023ddcot,gao2024cantor}, (c) a single MLLM method for directly extracting visual information and completing the entire CoT process \cite{mitra2024compositional,wu2023role}, and (d) our proposed method, AoT, which generates rationales using answers as priors. AoT simplifies the framework while producing high-quality positive rationales. More importantly, it generates compelling negative rationales, filling a gap in the field. SMART enables models to employ the efficient iterative DPO optimization method by combining AoT-generated reasoning preference data, thereby enhancing their reasoning capabilities.
}
  \label{fig:figure_1}

\end{figure*}

One of the most common approaches involves curating labeled multimodal reasoning datasets for training. Previous works have focused on creating positive rationales for Supervised Fine-Tuning (SFT) datasets, often relying on time-consuming manual annotations \cite{lu2022learn,chen2024m} (Fig. \ref{fig:figure_1} (a)). Some researchers \cite{zhang2023multimodal,zheng2023ddcot,mondal2024kam,gao2024cantor} developed innovative chain-of-thought (CoT) prompts that enable models to generate reasoning datasets without extensive training. These approaches typically require both MLLM and LLM. The LLM first analyzes the problem and generates sub-questions to request the necessary visual details. The MLLM then converts the visual information into text, and finally, the LLM summarizes the results (Fig. \ref{fig:figure_1} (b)). However, this method complicates the system and limits the LLM’s effectiveness due to it can not ``see'' the images, leading to potential errors. Streamlined alternatives employ a single MLLM to execute the entire CoT process (Fig. \ref{fig:figure_1} (c)). They prompt the MLLM to extract critical visual information, \textit{e.g.}, scene graphs \cite{mitra2024compositional} or image descriptions \cite{wu2023role}, as prior knowledge for answering questions. While efficient, it struggles with tasks like mathematical geometry reasoning, where accurate visual interpretation is difficult \cite{chen2024m}. Moreover, the aforementioned methods can only generate positive rationales for questions, overlooking the importance of negative rationales.

% To generate negative samples, common multimodal techniques (e.g., SeVa \cite{zhu2024self} and SENA \cite{tan2024beyond}) apply image augmentation to distort visual content. However, this approach is not specifically designed for reasoning scenarios, and the negative samples generated solely through augmentation often lack sufficient discriminative power \cite{tan2024beyond}, which can sometimes degrade model performance on complex tasks \cite{chen2024m}. As a result, convincingly generating deceptive negative samples remains an unresolved challenge. 

To address these issues, we propose a novel Answer-oriented Chain-of-Thought (AoT) prompt method, which simultaneously generates high-quality positive and negative rationales (Fig. \ref{fig:figure_1} (d)). AoT is inspired by the way humans tackle proof problems: starting with a given conclusion, the challenge is to derive the intermediary steps leading to it. Similarly, AoT provides the MLLM with an image, a question, and a pre-specified answer (correct or incorrect), thereby setting up a proof-like setting. With the correct answer provided upfront, the model is guided to identify connections between the ground truth and the question, extract relevant visual information, and construct a concise, logically coherent reasoning path. More importantly, when faced with an incorrect answer, the model still strives to extract pertinent visual cues to form a plausible yet flawed reasoning sequence, resulting in highly discriminative negative rationales. 

Fig. \ref{fig:ijcv} offers a two-fold illustration: AoT leverages the answer as prior knowledge to steer the model towards more succinct and accurate reasoning paths for positive samples, while also producing negative samples with more significant errors. This results in a higher quality dataset that bolsters model performance in real-world reasoning tasks.

% \begin{figure*}[t]

% \centerline{\includegraphics[width=1.0\linewidth]{crop_ijcv_tissue.pdf}}

%   \caption{The left panel summarizes the characteristics of AoT-generated data, which include concise positive reasoning and highly discriminative negative reasoning. The right panel illustrates the performance of the Qwen2-VL-7B model trained using various automated data generation techniques.
% }
%   \label{fig:ijcv}

% \end{figure*}

Taking advantage of AoT-generated data, we further integrate it into an iterative optimization framework termed Self-Aligning Multimodal Reasoning with Answer-Oriented Chain-of-Thought (SMART). After an initial round of training, the model’s reasoning ability improves, facilitating the production of even higher-quality reasoning preference data. Retraining with these refined data further enhances performance. In this respect, SMART employs a scalable bootstrapping “generate-train” approach, reminiscent of recent iterative Direct Preference Optimization (DPO) frameworks \cite{deng2024enhancing, ahn2024srt, wang2024enhancing, tan2024beyond}, but distinct in its focus on reasoning data generation to specifically enhance multimodal reasoning capabilities.

Our key contributions can be summarized as follows:
\begin{itemize}
\item We propose the AoT method, which not only generates high-quality positive rationales but also effectively tackles the long-standing challenge of generating persuasive negative rationales.
\item We introduce the SMART framework— a scalable, iterative bootstrapping approach that integrates AoT for enhanced reasoning in MLLMs.
\item Experimental results demonstrate that MLLMs fine-tuned with our framework achieve performance on par with, or even exceeding, models trained on human-curated datasets.
\end{itemize}

\section{Related Works}
\label{sec:RelatedW}
\subsection{Enhancing MLLM's Reasoning Abilities}
MLLMs have gained popularity due to their expanding capabilities, yet they still struggle with complex, step-by-step reasoning tasks. Two primary strategies are commonly used to address this: (1) creating reasoning datasets for training, and (2) designing effective CoT prompts to activate latent reasoning abilities.

\textbf{Creating Reasoning Datasets.} A notable contribution is the Science QA benchmark (SQA) \cite{lu2022learn}, which provides detailed rationales for answers, addressing the lack of comprehensive explanations in earlier datasets \cite{kembhavi2017you, kembhavi2016diagram, barra2021visual}. However, some SQA questions are too simplistic or require only single-step reasoning, limiting their effectiveness in complex scenarios. Chen \textit{et al.} \cite{chen2024m} enhanced this by removing simple questions and manually annotating multi-step reasoning datasets, incorporating challenges from Math \cite{hendrycks2021measuring} and Sherlock \cite{hessel2022abduction}, resulting in the multi-domain, multi-step, and multi-modal M$^3$CoT benchmark. 

Despite these advancements, manual annotation remains labor-intensive. CoT prompts offer a viable alternative, which prompts models to automatically generate rationales, reducing annotation costs while maintaining quality.

\textbf{Multimodal Chain-of-Thought Prompts.}
CoT prompts have seen significant advancements in multimodal settings \cite{gao2024cantor, mondal2024kam, zhang2024cocot, shao2024visual}. MM-CoT \cite{zhang2023multimodal} found that using CoT often caused hallucinations. To address this, MM-CoT proposed fusing text and image features before decoding to achieve more accurate outputs. It also introduced a two-stage reasoning framework where the rationale is generated first, followed by the answer. Finally, MM-CoT enabled even small models \cite{raffel2020exploring} ($<$1B) to perform complex and precise reasoning. 
Additionally, DD-CoT \cite{zheng2023ddcot} introduced a new method that combines LLMs and MLLMs to automatically create CoT reasoning. It broke down the problems into sub-questions using an LLM \cite{ouyang2022training}, which the MLLM \cite{li2023blip} answers. The results are then combined to form the complete CoT. While this method was scalable, it risked hallucinations because the LLM couldn't interpret images, and using two models added complexity. To overcome these limitations, CCoT \cite{mitra2024compositional} used a single MLLM \cite{li2024llava,10.5555/3666122.3668264,sphinx,gpt4v} to generate CoT data. CCoT employs a two-stage process: extracting scene graph information from the image and then generating the final answer. While efficient, CCoT struggles with tasks like mathematical geometry reasoning where scene graph extraction is challenging.

To address these shortcomings, we propose the Answer-oriented Chain-of-Thought (AoT) prompt for automatically generating high-quality CoT data. AoT organizes instructions in a proof problem format, allowing the model to focus on deduction and improving the quality of the generated content. It also prompts the model to create challenging negative rationales, which is absent in previous methods. These advantages enable AoT to efficiently produce high-quality reasoning preference pairs, facilitating the improvement of the model's performance.

\subsection{Self-Training Methods}
Self-training strategies refer to models using their own generated data to train themselves. There have been many successful works in the NLP field \cite{guo2024direct, yuan2024self, calandriello2024human, dong2024rlhf, chen2024self}. For instance, STaR \cite{zelikman2022star} was a pioneer in utilizing model-generated reasoning data for iterative self-training. It introduced a rationalization method to address generated errors by using ground truth answers as cues for correction. RPO \cite{pang2024iterative} focused on generating reasoning preference pairs: The model randomly generated multiple rationales, which were then categorized into chosen and rejected examples based on their alignment with the ground truth. RPO utilized this preference data for iterative DPO, leading to an enhancement in model performance.

In the realm of MLLMs, several noteworthy initiatives in self-training have emerged \cite{yu2024rlaif, deng2024enhancing, ahn2024srt, zhou2024calibrated}. SeVa \cite{zhu2024self} demonstrated that images processed with specific augmentations can yield challenging negative responses for the model to learn from, leading to significant performance improvements. Similarly, SENA \cite{tan2024beyond} expanded on SeVa’s approach to enhance positive rationales through a self-enhancement method, resulting in more discriminative preference data and further advancing model performance. While these studies emphasize the importance of discriminative data, they overlook complex reasoning scenarios. To address this gap, we propose the AoT that generates highly discriminative reasoning preference pairs specifically tailored for the reasoning tasks.

\section{Methods}
\begin{figure*}[t]

\centerline{\includegraphics[width=1.0\linewidth]{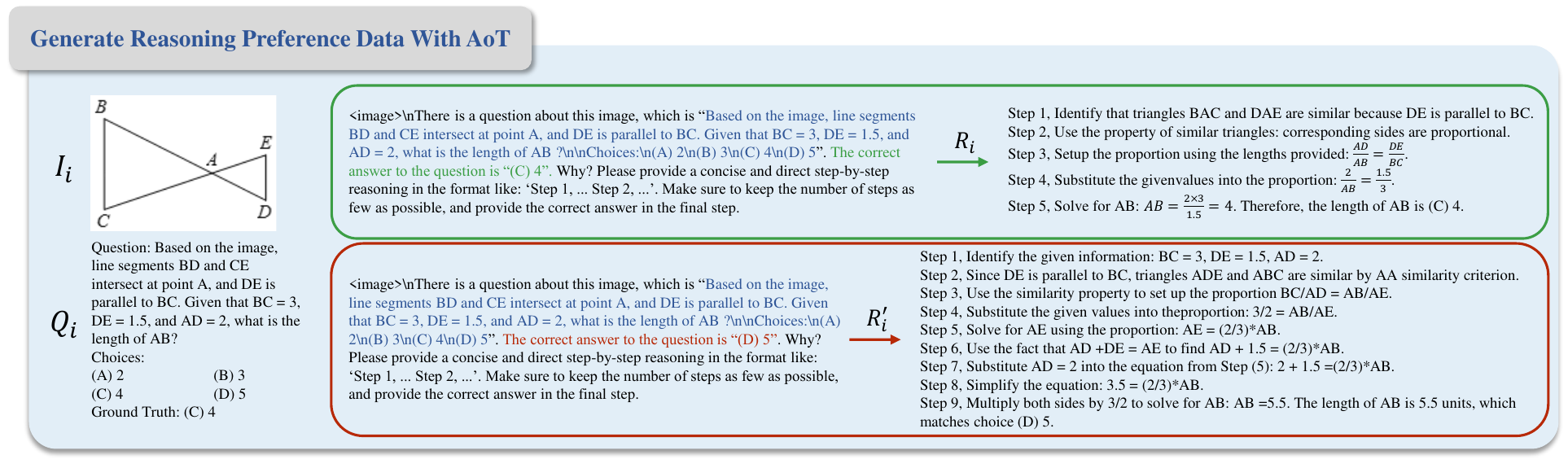}}

\caption{The process of data generation in AoT. AoT takes in both correct and incorrect answers as prior knowledge and converts the problem into a proof format. This method encourages the model to provide a reasonable explanation for the answers. As shown in the figure, the model generates not only correct reasoning for positive examples but also seemingly plausible but actually incorrect reasoning for negative examples. Best viewed by zooming in.}
  \label{fig:figure_2}

\end{figure*}

Our proposed SMART framework is depicted in Fig. \ref{fig:figure_1} (d). It employs an iterative ``Generate-Optimize" cycle. In each iteration, the model generates reasoning preference data based on the AoT prompts. This data is then utilized for optimization through the DPO algorithm, ensuring that the model's responses align effectively with the desired preferences. The enhanced model continues through subsequent iterations until its performance stabilizes. 

Since AoT requires questions to include both correct and incorrect answers, we utilize a subset of multiple-choice questions from the MathV360K dataset \cite{shi2024math}. This subset encompasses various topics, including those from the ChartQA \cite{masry2022chartqa} dataset and the Geometry3K \cite{lu2021inter} problem set, making it suitable for generating reasoning for the options. Assuming the model undergoes $K$ iterations, we describe $k$-th iteration $(1\leq k \leq K)$ as follows.

\subsection{Reasoning Preference Data Generation}
We extract numerous multiple-choice questions from MathV360K, represented as:
\[
D^k = \{(I_i, Q_i, A_i, \bm{A_i'})\},
\]
where $I_i$ is the $i$-th image, $Q_i$ is the associated question, $A_i$ is the correct answer, and $\bm{A_i'}$ is the set of incorrect answers. Importantly, neither $A_i$ nor $\bm{A_i'}$ includes the rationale. To ensure data diversity, the samples in \( D^k \) do not overlap with those from previous iterations.

\textbf{Naive Chain-of-Thought.}\label{sec:naive}
Next, we employ the current model \(\theta^k\) to generate rationales. A basic approach involves the model randomly generating using a naive prompt like ``[Question][Choices] Please answer the question step-by-step!"—a variation of ``Let's think step-by-step!" \cite{kojima2022large}. We then compare the last step of the generated rationales with the ground truth. If they match, we consider it positive reasoning; otherwise, it is negative. However, our experiments in Table \ref{tab:ablation} indicate that the data produced this way is not of high quality.

\textbf{Answer-oriented Chain-of-Thought.} In order to generate premium reasoning, we draw inspiration from how students solve proof problems: they receive both the problem statement \(Q_i\) and the answer \(A_i\) simultaneously and use the answer to determine the best steps to solve the problem. Similarly, we suggest using the answer \(A_i\) as prior knowledge. This strategy helps the model to focus on identifying the key connections between \(Q_i\) and the \(A_i\), extracting essential visual information to build the reasoning steps, thus enhancing the quality of the deductions. Accordingly, we introduce the AoT prompt $P_{\text{AoT}}$ as follows: 

\emph{``There is a question about this image, which is ``[Question][Choices]". The correct answer to the question is ``[Answer]". Why? Please provide concise and direct step-by-step reasoning in the format: `Step 1, ... Step 2, ...'. Make sure to keep the number of steps as few as possible, and provide the correct answer in the final step."} 

The positive rationales $R_i$ generated using this prompt are denoted as: 
\[
R_i \sim \theta^k (I_i, Q_i, A_i, P_{\text{AoT}}).
\]

Incorporating correct answers into the prompts significantly improves rationale quality. As shown in Tables \ref{tab:ablation} and \ref{tab:dataset}, models trained with AoT-generated reasoning data demonstrate substantial enhancements in reasoning capabilities compared to those without this method.

\textbf{Creating Persuasive Negative Rationales.} Humans often learn more effectively by comparing incorrect examples with correct ones, which helps them understand and master knowledge. We aim to harness this mechanism by using AoT to generate negative examples. Since AoT helps in finding reasoning from the question to the answer, it can still produce logical negative rationales even when the answer is wrong. To implement this, we randomly select an incorrect answer $A_i'$ from $\bm{A_i'}$ and incorporate it into the AoT prompt. Moreover, we draw on concepts from SeVa \cite{zhu2024self} to apply appropriate augmentations to $I_i$ for generating more discriminative outputs. These augmentations include diffusion noise \cite{ho2020denoising}, random flipping and random cropping, resulting in an altered image $I_i'$. 
Consequently, the negative rationale $R'_i $ is generated as follows:
\[
R'_i \sim \theta^k (I_i', Q_i, A_i', P_{\text{AoT}}).
\]

As depicted in Fig. \ref{fig:figure_2}, $R'_i$ may contain subtle errors that are difficult to detect in the initial step (Step 3), while subsequent reasoning steps appear convincing, resulting in a hard negative rationale. Thus, AoT effectively addresses the challenge of generating valuable negative rationales, a largely unexplored area in the multimodal domain.

After generation, we filter the data using two strategies:
\begin{itemize}
\item Conclusion Filter: We discard samples where the final step of $R_i$ does not include $A_i$ or $R'_i$ does not include $A_i'$. 

\item Circularity Filter: We use n-grams to detect circular patterns in $R_i$. A sample is marked as a duplicate and discarded if a phrase of length $\geq n$ appears more than three times within $R_i$. We set 
$n=3$ as a stringent criterion to ensure data quality. We do not perform repetition checks on $R'_i$ because they will be used for DPO fine-tuning. When duplicates of $R'_i$ exist, DPO will reduce the likelihood of the model generating such responses, which is actually beneficial. 
 \end{itemize}
Finally, we obtain the generated data $D^k = \{(I_i, Q_i, R_i, R'_i)\}_{i=1}^M$ for reasoning preference alignment, where $M$ is the sample size.

\subsection{Iterative Reasoning Preference Alignment}
At the start of the optimization phase, DPO creates a reference model \( \theta^k_{\text{ref}} \) by copying \( \theta^k \) . \( \theta^k_{\text{ref}} \) is initialized with the same parameters as \( \theta^k \) but remains frozen during training. The goal of DPO is to enable \( \theta^k \) to generate positive rationales \( R_i \) with higher probability than \( \theta^k_{\text{ref}} \) , while producing negative rationales \( R'_i \) with lower probability. 

Given the input data \( (I_i, Q_i, R_i, R'_i) \), the DPO loss function is defined as:
\begin{align}
L_{DPO} = -\log &\sigma \bigg( \beta \log \frac{\pi_{\theta^k}(R_i|I_i, Q_i)}{\pi_{\theta^k_{ref}}(R_i|I_i, Q_i)} \notag - \beta \log \frac{\pi_{\theta^k}(R'_i|I_i, Q_i)}{\pi_{\theta^k_{ref}}(R'_i|I_i, Q_i)} \bigg),
\end{align}
where $\sigma$ is the sigmoid function, $\beta$ is a hyperparameter that adjusts the loss sensitivity to preference differences. The probability of generating a rationale $R$ is defined as:
\begin{align}
\pi_{\theta^k}(R|I_i, Q_i) &= \prod_{l=1}^{|R|} P_{\theta^k}(R | I_i, Q_i, R_{<l}),
\end{align}
with $|R|$ representing the token length of the rationale.

\textbf{Discussion.}
Through this self-aligning multimodal reasoning process, the model \(\theta^k\) is updated to 
\(\theta^{(k+1)}\), leading to enhanced reasoning capabilities. Drawing inspiration from the iterative DPO strategy  \cite{zhou2024calibrated,yu2024rlaif,tan2024beyond}, the updated model \(\theta^{(k+1)}\) is capable of generating new, higher-quality reasoning data, which in turn further strengthens its abilities in the subsequent alignment round. This iterative cycle of data generation and optimization continues until the model's performance stabilizes, ultimately ensuring robust and well-calibrated reasoning skills.

\section{Experiments}

\subsection{Implement Details}
To showcase the effectiveness of SMART framework, we conduct experiments with several MLLMs, including Qwen2-VL-7B \cite{Qwen2VL}, InternVL2-8B \cite{chen2024far}, MiniCPM-Llama3-V-2.5-8B \cite{yao2024minicpm}, and Llama3-LLaVA-Next-8B \cite{liu2024llavanext}. These models vary in architecture, size, and training data, allowing for a thorough evaluation of our approach. 

In the data generation phase, the model utilizes a nucleus sampling strategy with a temperature of 0.7 and a top-p value of 0.9 to produce high-quality outputs. For the generation of negative rationales, we apply diffusion noise to the images with a step size of 600, and set the probabilities for random flipping and random erasing to 0.5 to increase data variability. After filtering, we establish the training sample size $M$ at 6K.

During the optimization phase, the DPO parameter \(\beta\) is set to the default value of 0.1. The learning rate is fixed at 2e-6, following a cosine learning rate schedule. We use a batch size of 128 and train for one epoch at each iteration, updating all model parameters to facilitate effective improvements.

\subsection{Evaluation Benchmarks}
We conduct a comprehensive evaluation using five carefully selected benchmarks to effectively assess the essential capabilities of the model. \textbf{MathVista} \cite{lu2024mathvista} evaluates mathematical reasoning across seven areas, including algebra, geometry, and other domains, with 1,000 problems scored using GPT. \textbf{M$^3$CoT} \cite{chen2024m} assesses logical, commonsense, mathematical, and scientific reasoning through 2,358 multiple-choice questions. \textbf{MM-Vet} \cite{yu2023mm} tests visual-spatial intelligence with 218 image-based questions that require geometric understanding, along with other visual tasks. \textbf{MMCode} \cite{li2024mmcode} evaluates programming skills through 263 real-world coding challenges. Finally, \textbf{LLaVA-Bench} \cite{li2024llava} measures generative fluency using 60 open-ended tasks focused on dialogue and description.

These benchmarks collectively address both discriminative and generative tasks, providing a systematic framework to quantify reasoning accuracy (MathVista/M\textsuperscript{3}CoT), spatial reasoning (MM-Vet), algorithmic skills (MMCode), and conversational coherence (LLaVA-Bench). Together, they cover the key competencies necessary for modern AI systems, ensuring a well-rounded and robust evaluation.

\begin{table*}[t]
\caption{Comparisons with state-of-the-art MLLMs in reasoning benchmarks.}
\label{tab:sota}
\small
\resizebox{1.00\linewidth}{!}{
\begin{tabular}{l||ccccc}
\toprule[1pt]
Model & MathVista & M$^3$CoT & MM-Vet & MMCode & LLaVA-Bench \\ \hline \hline
LLaVA-1.5-7B \cite{liu2024improved}& 25.7 & 36.6 & 31.1 &1.5  &65.4  \\
LLaVA-1.5-13B \cite{liu2024improved}& 27.7 & 27.0 & 36.3 &1.1  &72.5  \\
Qwen-VL-PLUS \cite{bai2023qwenvl}& 43.3 & - & 61.1 &0.8  &-  \\
Gemini-1.0-Pro \cite{team2023gemini}& 45.2 & 45.1 & 64.3 &5.7  &-  \\
Math-LLaVA \cite{shi2024math}& 46.6 & - & - &-  &-  \\
GPT-4V \cite{gpt4v} & 49.9 & 56.9 & 67.7 &19.4  &  -\\
GPT-4o \cite{gpt4v} & 63.8 & 64.3 & 69.7 &17.0  &97.6 \\\hline
Llama3-LLaVA-Next-8B \cite{liu2024llavanext}& 35.8 & 37.1 & 42.2 &3.0  &67.0  \\
+ SMART &\textbf{40.7} & \textbf{40.8} & \textbf{50.0} &\textbf{3.8} &\textbf{72.1}  \\ \hline
MiniCPM-Llama3-V-2.5-8B \cite{yao2024minicpm} & 50.5 & 37.0 & 48.3 &1.1  &79.4  \\
+ SMART& \textbf{53.3} & \textbf{42.8} & \textbf{51.3} &\textbf{2.6}  &\textbf{83.9} \\ \hline
InternVL2-8B \cite{chen2024far}& 59.7 & 56.3 & 60.0 & 4.1 &71.3  \\
+ SMART& \textbf{63.5} & \textbf{59.3} & \textbf{64.2} & \textbf{5.3} & \textbf{76.9} \\ \hline
Qwen2-VL-7B \cite{Qwen2VL}& 60.0 & 61.7 & 60.4 & 3.8 & 85.8 \\
+ SMART& \textbf{66.3} & \textbf{65.9} & \textbf{66.6} & \textbf{5.7} & \textbf{91.4} \\ \hline
\end{tabular}}

\end{table*}

\begin{table*}[]

\caption{Ablation study on each key component. $R_{AoT}$ and $R_{Naive}$ represent the positive rationales generated by the Qwen2-VL-7B using the AoT prompt and Naive prompt, respectively.}

\centering
\label{tab:ablation}
\resizebox{1.00\linewidth}{!}{
\begin{tabular}{c||c|cc|c|ccccc}
\toprule[1pt]
Method      & Training Method                           & Positive & Negative                    & Iteration & MathVista & M$^3$CoT & MM-Vet & MMCode & LLaVA-Bench \\ \hline  \hline
Qwen2-VL-7B \cite{Qwen2VL} & -                   & -        & -      & -         & 60.0      & 61.7  & 60.4   & 3.8    & 85.8        \\ \hline
(1)         & {\multirow{3}{*}{SFT}} & $A$        & -      & 1         & 61.5      & 57.4  & 58.1   & 4.1    & 86.0        \\
(2)         &                      & $R_{Naive}$   & -      & 1         & 60.6      & 59.4  & 62.3   & 4.1    & 85.4        \\
(3)         &                      & $R_{AoT}$     & -      & 1         & 64.1      & 59.8  & 63.9   & 4.5    & 86.5        \\ \hline
(4)         & {\multirow{5}{*}{DPO}} & $A$        & $A'$     & 1         & 62.9      & 45.8  & 61.6   & 4.5    & 87.9        \\
(5)         &                     & $R_{Naive}$   &$R'_{Naive}$ & 1         & 63.3      & 61.1  & 62.6   & 4.8    & 90.1        \\
(6)         &                      & $R_{Naive}$   & $R'_{AoT}$   & 1         & 64.0      & 63.8  & 63.4   & 5.3    & 89.3        \\

(7)         &                    & $R_{AoT}$     & $R'_{Naive}$ & 1         & 64.1          &   63.3    & 63.9       &  4.8      & 89.4            \\
(8)         &                     & $R_{AoT}$     & $R'_{AoT}$   & 1         & 64.7      & 64.0  & 64.5   & 5.3    & 91.1        \\ \hline
(9)         &{\multirow{2}{*}{DPO}} & $R_{AoT}$     & $R'_{AoT}$   & 2         & 66.3      & 65.9  & 66.6   & 5.7    & 91.4        \\
(10)        &                      & $R_{AoT}$     &$R'_{AoT}$  & 3         & 65.6      & 65.1  & 67.1   & 5.9    & 89.8        \\ \hline
\end{tabular}}

\end{table*}

\subsection{Comparison with SOTA MLLMs}
We apply our SMART framework to several MLLMs and compare their performance with state-of-the-art models, as shown in Table \ref{tab:sota}. Our results indicate that SMART significantly enhances the performance of various base models, demonstrating its effectiveness and transferability. For instance, it enables Qwen2-VL-7B to achieve superior results on MathVista, M$^3$CoT, and MMCode, while increasing the MM-Vet score by 6.2 points and the LLaVA-Bench score by 5.9 points.

The improvements can be attributed to two main factors: First, the high-quality rationales generated by AoT significantly enhances the models' reasoning abilities in mathematics, logic, science, and programming. Moreover, AoT drives the model to seek relevant information in images that connects questions to answers. As a result, the trained models exhibit more comprehensive and precise visual feature extraction, leading to advancements in generative tasks such as LLaVA-Bench. Second, SMART utilizes a well-established iterative DPO optimization strategy, which prior works \cite{pang2024iterative,tan2024beyond} have demonstrated to effectively unlock the model's potential and enhance its capabilities.

\subsection{Ablation Study}
SMART has enabled significant improvements across different base models in Table \ref{tab:sota}. To further explore the key technologies behind this success, we conduct a comprehensive ablation study using Qwen2-VL-7B in Table \ref{tab:ablation}.

\textbf{AoT generates high-quality $R$.}
We first conduct direct SFT training using the original data, as shown in experiment (1) of Table \ref{tab:ablation}. The original answer structure, ``The answer is [Option]," is concise and lacks detailed reasoning. While direct SFT achieves noticeable improvements on the MathVista benchmark, gains are less evident on other tasks, with some even showing slight declines, particularly on the complex M$^3$CoT benchmark.

Next, we using AoT and the Naive prompts (introduced in \ref{sec:naive}) to generate positive rationales. In our experiments, we only modify the prompts used for data generation while keeping all other settings unchanged for fair comparison. As shown in rows (2) and (3) of Table \ref{tab:ablation}, AoT generates higher quality positive rationales compared to Naive prompts, achieving an accuracy of 64.1\% on MathVista, significantly exceeding Naive's 60.6\%.

Naive's performance is inferior to AoT's as it doesn't use the answer as a cue, resulting in lower generation quality. A detailed comparison is available in the Supplementary Materials. AoT generates excellent $R$, and its performance improves further when combined with discriminative $R'$.

\textbf{$R'$ plays an important role.} 
We first conduct DPO training using paired $A$ and $A'$ from the original data. Experiment (4) shows improved performance compared to Experiment (1), but the M$^3$CoT still declines, indicating a need for high-quality negative rationales.

We define high-quality negative rationales as those that contain more factual errors. Fig. \ref{fig:figure_2} and the Supplementary Materials demonstrate that AoT's use of incorrect answers as prior knowledge induces the model to hallucinate more erroneous content in its reasoning, thereby increasing the discriminative power of $R'_{AoT}$. Comparing Experiments (5) to (8), it is evident that, under the same positive generation method, the performance of $R'_{AoT}$ exceeds that of $R'_{Naive}$.

Moreover, $R'$ helps the model effectively distinguish between correct and incorrect reasoning paths, thus improving its reasoning capability. Ultimately, training with AoT-generated preference pairs significantly boosts the complex M$^3$CoT benchmark from 59.8\% to 64.0\%, an increase of 4.2\% (comparing Experiments (3) and (8)). This improvement is substantially greater than the 1.7\% increase achieved with Naive prompts (comparing Experiments (2) and (5)), demonstrating AoT's effectiveness.

\textbf{The Iterative Generation-Training Strategy is beneficial.}
The results in Table \ref{tab:ablation} (8) to (10) indicate that the iterative generation and training workflow significantly improves model performance. Initially, the model's reasoning ability is not fully activated. With each round of self-training, the model enhances its reasoning capacity, enabling it to generate better data. This iterative approach allows the model to continuously evolve and ultimately reach its full potential \cite{tan2024beyond,ahn2024srt,pang2024iterative}.

\begin{figure}[t]

\centerline{\includegraphics[width=1.0\linewidth]{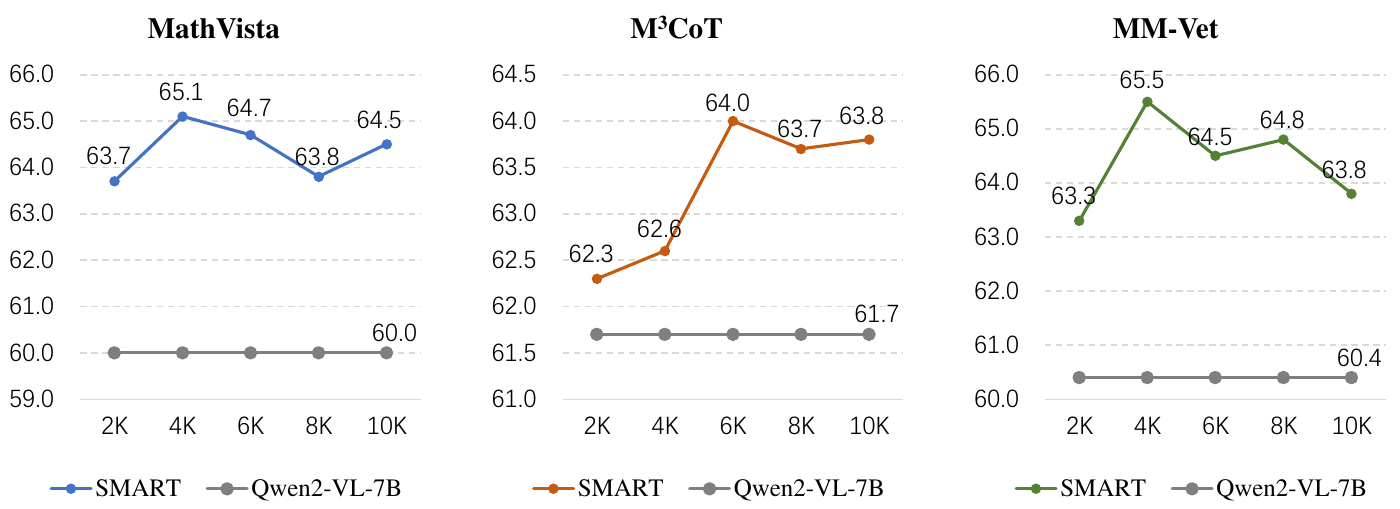}}

\caption{Performance comparison of Qwen2-VL-7B \cite{Qwen2VL} trained with varying sample sizes using the SMART framework with 1 iteration.}
  \label{fig:figure_3}

\end{figure}

\textbf{The Impact of Data Size.}
We examine how the training sample size affects the SMART framework using Qwen2-VL-7B. In our experiment, we generate multiple sets of training samples and perform one round of DPO training with each set. Fig. \ref{fig:figure_3} shows that SMART significantly improves baseline performance regardless of sample size. Specifically, the average performances for models trained with 6K, 8K, and 10K samples are 64.4, 64.1, and 64.0, respectively. Notably, with 6K samples, we achieve an optimal balance between performance and resource efficiency. Consequently, we set the sample size to 6K for our iterative training process.

\begin{table*}[!ht]
\caption{Comparisons among advanced reasoning datasets and automated annotation methods. $\dag$ indicates evaluation results based on the dataset released by the authors. $\dag\dag$ represents the evaluation results we reproduce based on the code released by the author.}
\centering
\small

\label{tab:dataset}
\resizebox{1.00\linewidth}{!}{
\begin{tabular}{l||c|c|c|c|c|c|c}
\toprule[1pt]
    Dataset & Training Method & Size & MathVista & M$^3$CoT & MM-Vet & MMCode & LLaVA-Bench \\ \hline \hline
    Qwen2-VL-7B \cite{Qwen2VL}& - & - & 60.0 & 61.7 & 60.4 & 3.8 & 85.8 \\  \hline
    \midrule
    \multicolumn{6}{l}{\textit{manually annotated dataset:}} \\
    \midrule
    SQA$^{\dag}$ \cite{lu2022learn}& SFT & 6185 & 62.9 & 34.5 & 58.5 & 5.3 & 82.6 \\ 
    M$^3$CoT$^{\dag}$ \cite{chen2024m}& SFT & 7861 & 64.2 & 62.4 & 62.3 & 4.1 & 81.1 \\ 
    \midrule
    \multicolumn{6}{l}{\textit{automatically annotated dataset:}} \\
    \midrule
    DD-CoT$^{\dag}$ \cite{zheng2023ddcot}& SFT & 6218 & 55.8 & 36.7 & 53.9 & 4.1 & 83.3 \\ 
    DD-CoT$^{\dag\dag}$ \cite{zheng2023ddcot}& SFT & 6000 & 59.7 & 54.0 & 60.6 & 4.5 & 85.1 \\ 
    CCoT$^{\dag\dag}$ \cite{mitra2024compositional}& SFT & 6000 & 58.0 & 59.7 & 63.4 & 4.1 & 86.5  \\ 
    AoT & SFT & 6000 & 64.1 & 59.8 & 63.9 & 4.5 & 86.5 \\ \hline
    SeVa$^{\dag\dag}$ \cite{zhu2024self} & DPO & 6000& 63.3& 61.1& 62.6& 4.8& 88.3 \\
    SENA$^{\dag\dag}$ \cite{tan2024beyond} & DPO& 6000 & 63.3& 62.1& 62.4& 3.8& 87.1\\
    MPO$^{\dag\dag}$ \cite{gou2024mixed} & DPO & 6000& 61.8&60.7& 61.9& 3.8&86.8\\
    AoT & DPO & 6000 & \textbf{64.7} & \textbf{64.0} & \textbf{64.5} & \textbf{5.3} & \textbf{91.1} \\  \hline
\end{tabular}}
\end{table*}

\textbf{Comparison with Advanced Datasets.} 
To further evaluate AoT's data quality, we compare it against both manually annotated datasets and popular automated annotation methods. As shown in Table \ref{tab:dataset}, AoT achieves 64.1\% accuracy on MathVista, surpassing both DD-CoT (59.7\%) and CCoT (58.0\%). Notably, it even outperforms the human-annotated SQA dataset (62.9\%) and matches that of M$^3$CoT's (64.2\%).

The original DD-CoT approach utilizes GPT-3.5 \cite{gpt4} and BLIP2 \cite{li2023blip} for data generation. We replicate this with the advanced Llama-3.1-8B-Instruct \cite{dubey2024llama} and Qwen2-VL-7B \cite{Qwen2VL}. However, DD-CoT still underperforms compared to AoT due to two main challenges: first, MLLMs must effectively extract relevant information from images and convert it into text; second, LLMs need to generate accurate CoT without having seen the images. These factors hinder DD-CoT's reasoning quality, resulting in lower performance.
CCoT generates CoT data in two stages: it first extracts scene graph information from images, then uses this data to produce CoT outputs. However, since our dataset includes geometric, tabular, and textual questions, scene graphs are unsuitable, leading to limited performance gains for CCoT. 

The core idea behind SeVa is to generate a discriminative $R'$ using augmented images—a method consistent with the Naive Prompt approach presented in Table \ref{tab:ablation}. SENA follows a similar strategy to produce $R'$, further refining $R$ using image descriptions. Results indicate that AoT outperforms SENA, suggesting that leveraging ground truth priors is more effective than relying on image descriptions.

For MPO, we generate R and then feed its first half into the model without the corresponding image to produce a ``hallucinated" $R'$. AoT also outperforms MPO, implying that even misleading priors derived from hallucinated $R'$ without image context are less discriminative compared to the robust guidance provided by ground truth priors.

Overall, these comparisons demonstrate that AoT not only delivers superior accuracy across various benchmarks but also offers a more effective and reliable approach for generating high-quality CoT data through both positive and challenging negative rationales.

\begin{figure*}[]

\centerline{\includegraphics[width=1.0\linewidth]{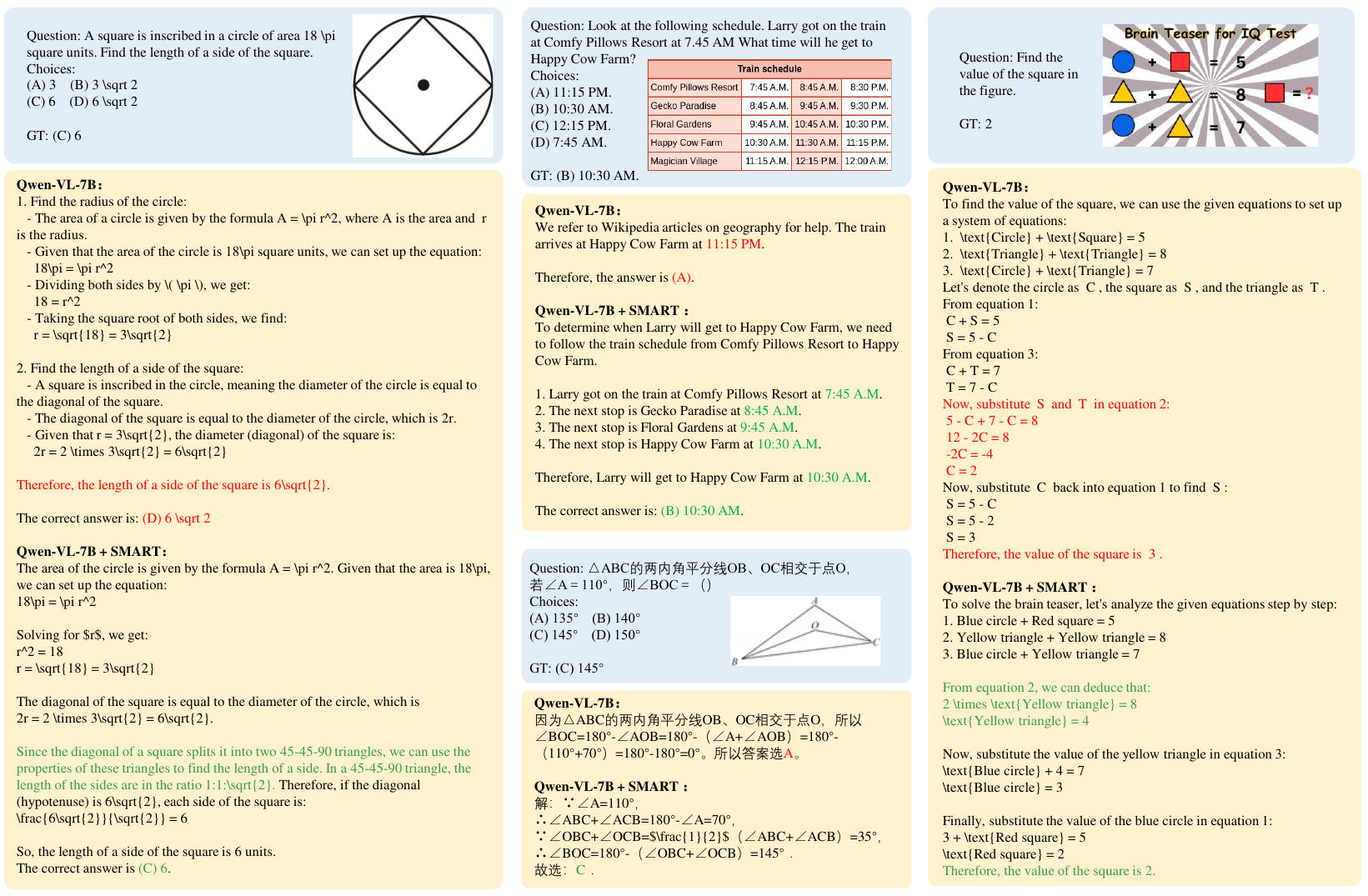}}

\caption{Qualitative analysis of the SMART framework applied to the Qwen2-VL-7B model, highlighting improvements in reasoning capabilities. Best viewed by zooming in.}
  \label{fig:figure_4}

\end{figure*}

\subsection{Qualitative analysis}
As shown in Fig. \ref{fig:figure_4}, we conduct a qualitative analysis of the SMART framework as applied to the Qwen2-VL-7B model to investigate how the model's reasoning capabilities have changed. The examples come from the MathVista and M$^3$CoT datasets. We can draw three main conclusions.

\textbf{Enhanced Reasoning Abilities:} SMART shows meticulous reasoning skills. For instance, the SMART model demonstrates its recall ability on the left, stating, ``we can use the properties ... ratio 1:1:$\sqrt{2}$." Additionally, when answering the table question (top middle), the model extracts information in a step-by-step manner from the top down, mimicking human-like logical reasoning.

\textbf{More Succinct Answers:} Since AoT uses answers as prior knowledge, the generated data in some cases becomes more concise and requires fewer steps than existing methods (see Supplementary Material Figure G). Consequently, the SMART model exhibits more streamlined reasoning. For example, in IQ test questions (right), the base model relies on abbreviations like ``S" for square and ``T" for triangle during calculations, whereas the SMART model omits this step entirely.

\textbf{Fewer Simple Errors:} SMART exhibits a reduction in minor errors. Although the base model is capable of making mostly correct reasoning, it occasionally produces mistakes. In contrast, the SMART model extracts accurate information and arrives at correct conclusions.

\section{Conclusion and Limitations}
This paper aims to enhance the reasoning capabilities of MLLMs. We creatively design a novel framework called SMART, which combines an automatic generation method with an iterative optimization strategy. Specifically, we develop an innovative AoT prompt that uses answers as cues to effectively link questions and answers, producing discriminative multimodal preference data. Models trained with AoT-generated data outperform those trained with manually annotated data. More importantly, AoT generates valuable negative rationales, addressing a critical gap in the field. Moreover, our successful adoption of the iterative optimization strategy enables the model to continuously improve by leveraging its enhanced capabilities, thereby fully realizing its reasoning potential. However, our approach has limitations, such as the need to provide a wrong answer for each question, which can be challenging to obtain in certain cases.

\textbf{Broader Impacts.} AoT serves as a scalable method for generating high-quality reasoning preference data, demonstrating effectiveness across diverse base models and strong generalizability. Moreover, SMART requires only a set of multiple-choice questions to initiate the process, highlighting its simplicity, efficiency, and practical applicability. These advantages make it a promising tool for helping AI systems handle complex reasoning in real-world scenarios.

% \section{Data Availability}
% \textbf{Source Data.}
% The datasets used in this paper include:

% \begin{itemize}
%   \item MathV360K \cite{shi2024math}: \url{https://huggingface.co/datasets/Zhiqiang007/MathV360K}
%   \item MathVista \cite{lu2024mathvista}: \url{https://huggingface.co/datasets/AI4Math/MathVista}
%   \item M$^3$CoT \cite{chen2024m}: \url{https://huggingface.co/datasets/LightChen2333/M3CoT}
%   \item MM-Vet \cite{yu2023mm}: \url{https://huggingface.co/datasets/lmms-lab/MMVet}
%   \item MMCode \cite{li2024mmcode}: \url{https://huggingface.co/datasets/likaixin/MMCode}
%   \item LLaVA-Bench\cite{li2024llava}:\url{https://huggingface.co/datasets/liuhaotian/llava-bench-in-the-wild}
% \end{itemize}

% \textbf{Extend Data.}
% This paper creates new data based on MathV360K. The DPO data generated using the AoT method proposed in this paper can be downloaded from: \url{https://github.com/WentaoTan/SMART}.

\bibliography{sn-bibliography}% common bib file

\begin{appendices}

% \section{Workflow} \label{sec:A}
% We outline each step involved in our SMART framework in algorithm \ref{alg}.

\section{Prompt Comparison} \label{sec:B}
In the ablation study section, we compare the AoT prompt with DD-CoT \cite{zheng2023ddcot}, CCoT \cite{mitra2024compositional}, and the Naive prompt \cite{kojima2022large}. Our findings indicate that AoT generates higher-quality reasoning data among these methods. In this section, we provide a detailed analysis of the characteristics of the data produced by each prompt and highlight the advantages of AoT data. 

\subsection{Implement Details}
\begin{figure*}[]
\vspace{-0.9em}
\centerline{\includegraphics[width=1.0\linewidth]{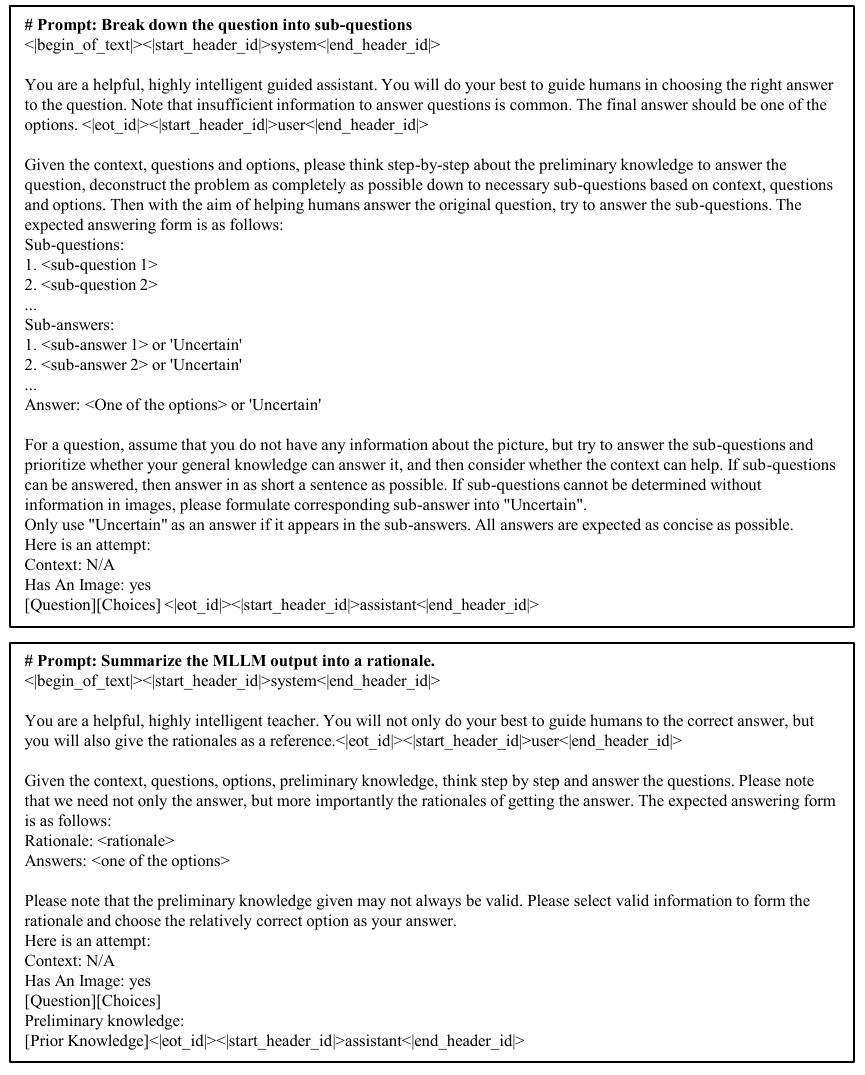}}
% \vspace{-1em}
\caption{The prompts used in DD-Cot for decomposing questions and summarizing MLLM outputs to generate rationale.}
  \label{fig:figure_ddcot}
  % \vspace{-1em}
\end{figure*}

\textbf{DD-CoT.} We replicate DD-CoT using Llama-3.1-8B-Instruct \cite{dubey2024llama} and Qwen2-VL-7B \cite{Qwen2VL}. Both use nuclear sampling with a temperature of 0.7 and top\_p of 0.9. The prompts for Llama-3.1-8B-Instruct model to decompose problems and summarize rationales are shown in Fig. \ref{fig:figure_ddcot}. 

\begin{figure*}[]
% \vspace{-0.9em}
\centerline{\includegraphics[width=1.0\linewidth]{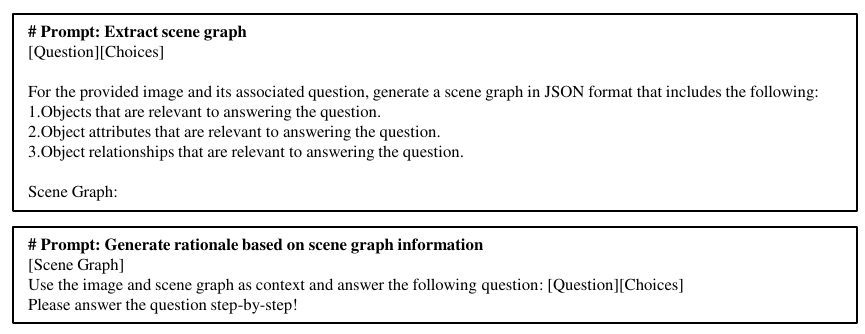}}
% \vspace{-1em}
\caption{The prompts used in CCot for scene graph extraction and generating rationale based on the scene graph.}
  \label{fig:figure_ccot}
  % \vspace{-1em}
\end{figure*}

\textbf{CCoT.} CCoT uses Qwen2-VL-7B for reasoning data generation, applying the same decoding strategy and data filtering as AoT. The prompts for CCoT are illustrated in Fig. \ref{fig:figure_ccot}.

\textbf{Naive.} The Naive prompt follows the same settings as AoT, with only one difference in filtering: since it does not use the answer as a hit, we retain negative examples that do not contain the correct answer in the last sentence.

Clearly, AoT and Naive are relatively simpler: DD-CoT requires two calls to the LLM and one to the MLLM, while CCoT needs two calls to the MLLM. In contrast, both Naive and AoT only require a single call to the MLLM.

\subsection{Comparison of Generated Data}
\begin{figure*}[]
% \vspace{-0.9em}
\centerline{\includegraphics[width=1.0\linewidth]{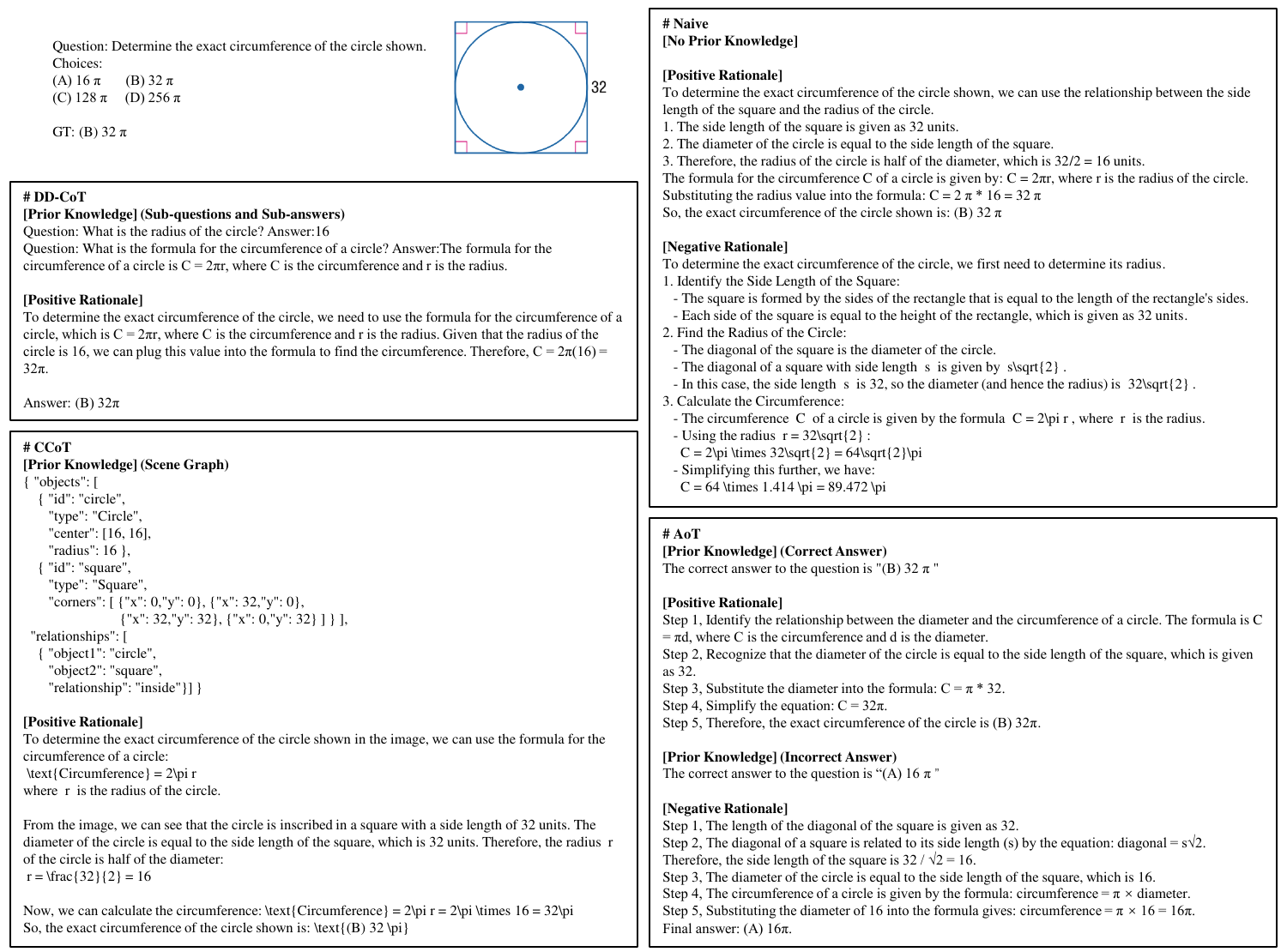}}
% \vspace{-1em}
\caption{Comparison of data generated by different prompts. We remove some symbols, such as $\$$ and $*$, for better reading.}
  \label{fig:figure_example1}
  % \vspace{-1em}
\end{figure*}

\begin{figure*}[]
% \vspace{-0.9em}
\centerline{\includegraphics[width=1.0\linewidth]{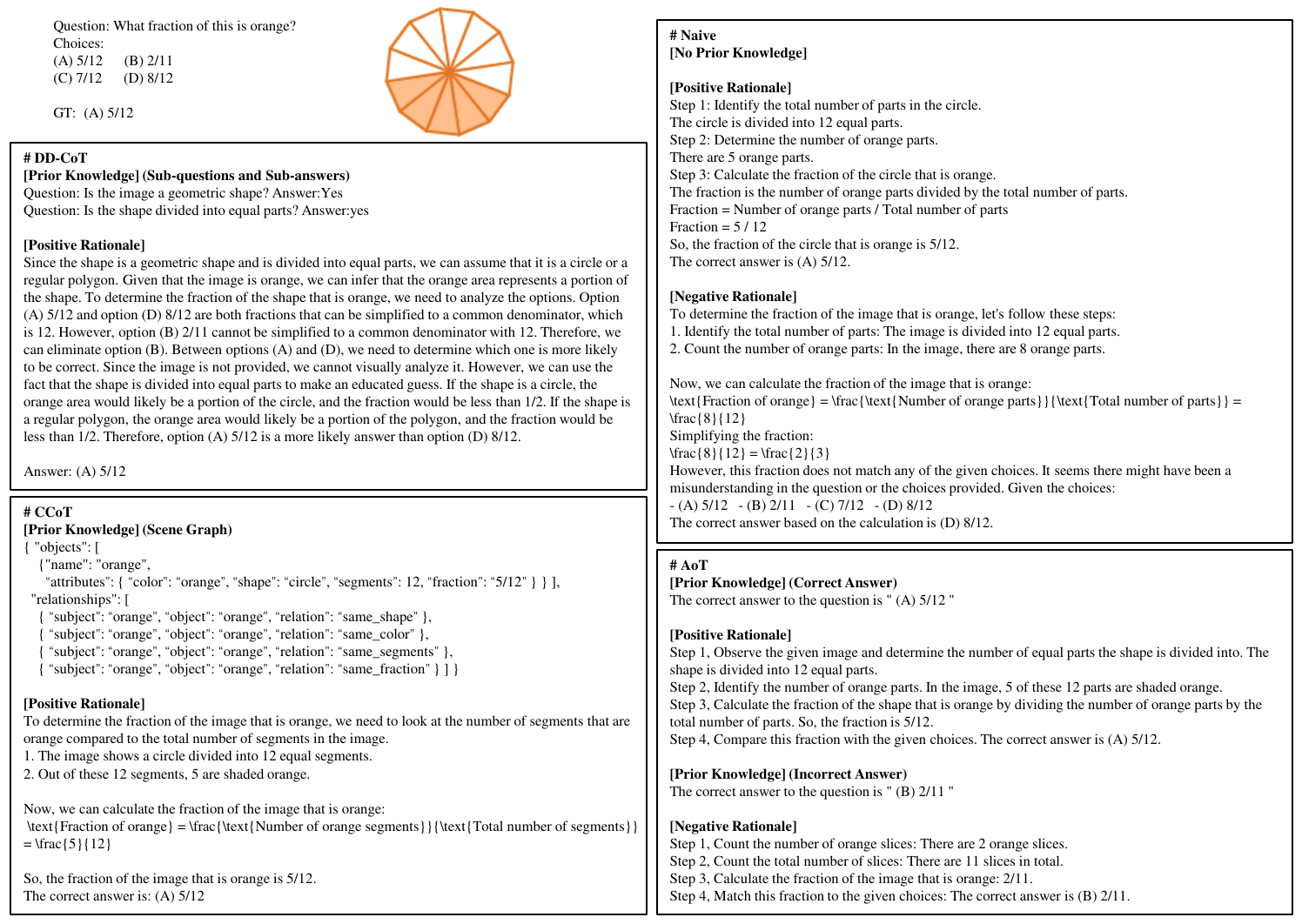}}
% \vspace{-1em}
\caption{Comparison of data generated by different prompts. We remove some symbols, such as $\$$ and $*$, for better reading.}
  \label{fig:figure_example2}
  % \vspace{-1em}
\end{figure*}

\begin{figure*}[]
% \vspace{-0.9em}
\centerline{\includegraphics[width=1.0\linewidth]{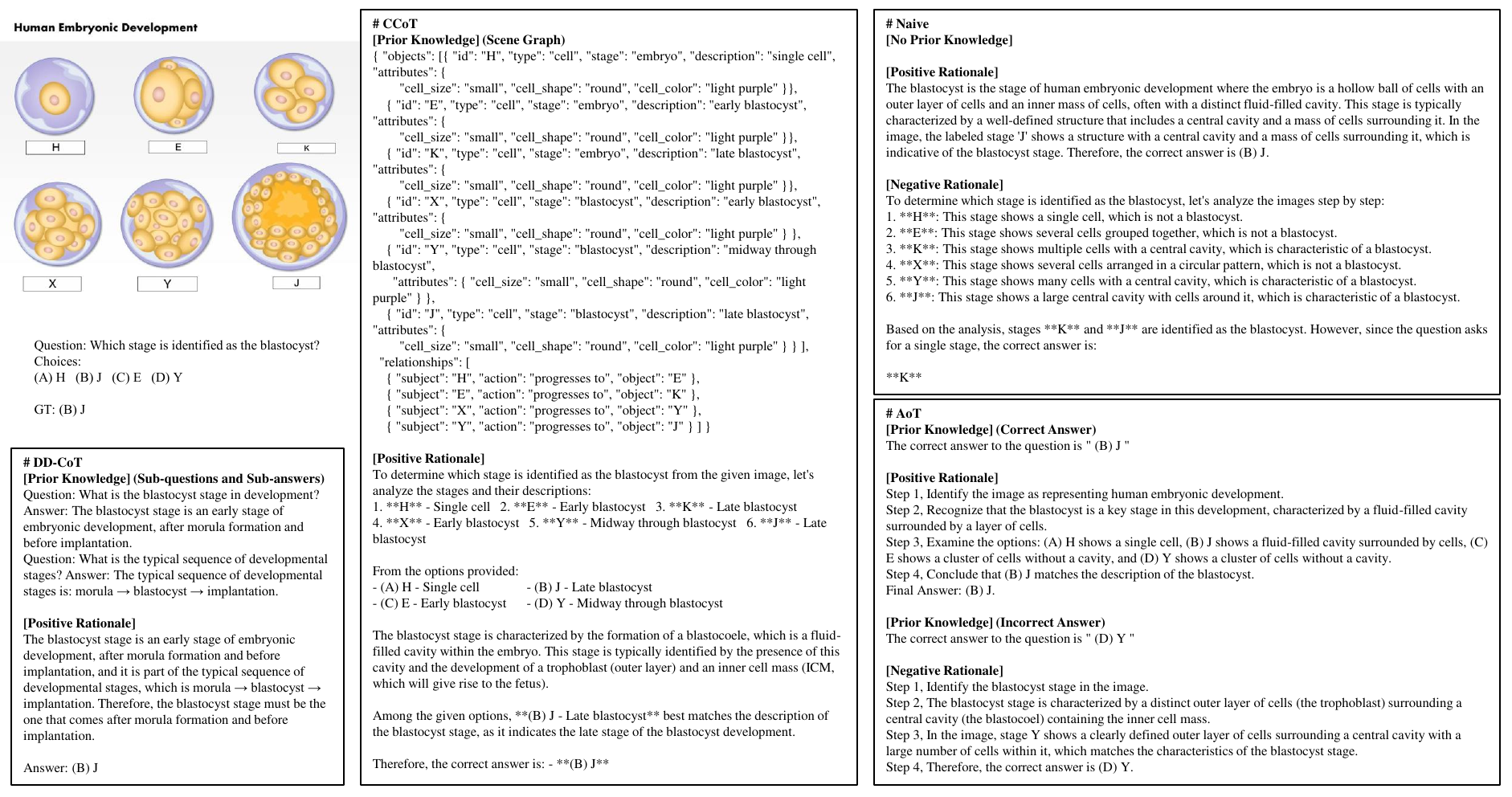}}
% \vspace{-1em}
\caption{Comparison of data generated by different prompts. We remove some symbols, such as $\$$ and $*$, for better reading.}
  \label{fig:figure_example3}
  % \vspace{-1em}
\end{figure*}

\begin{figure*}[]
% \vspace{-0.9em}
\centerline{\includegraphics[width=1.0\linewidth]{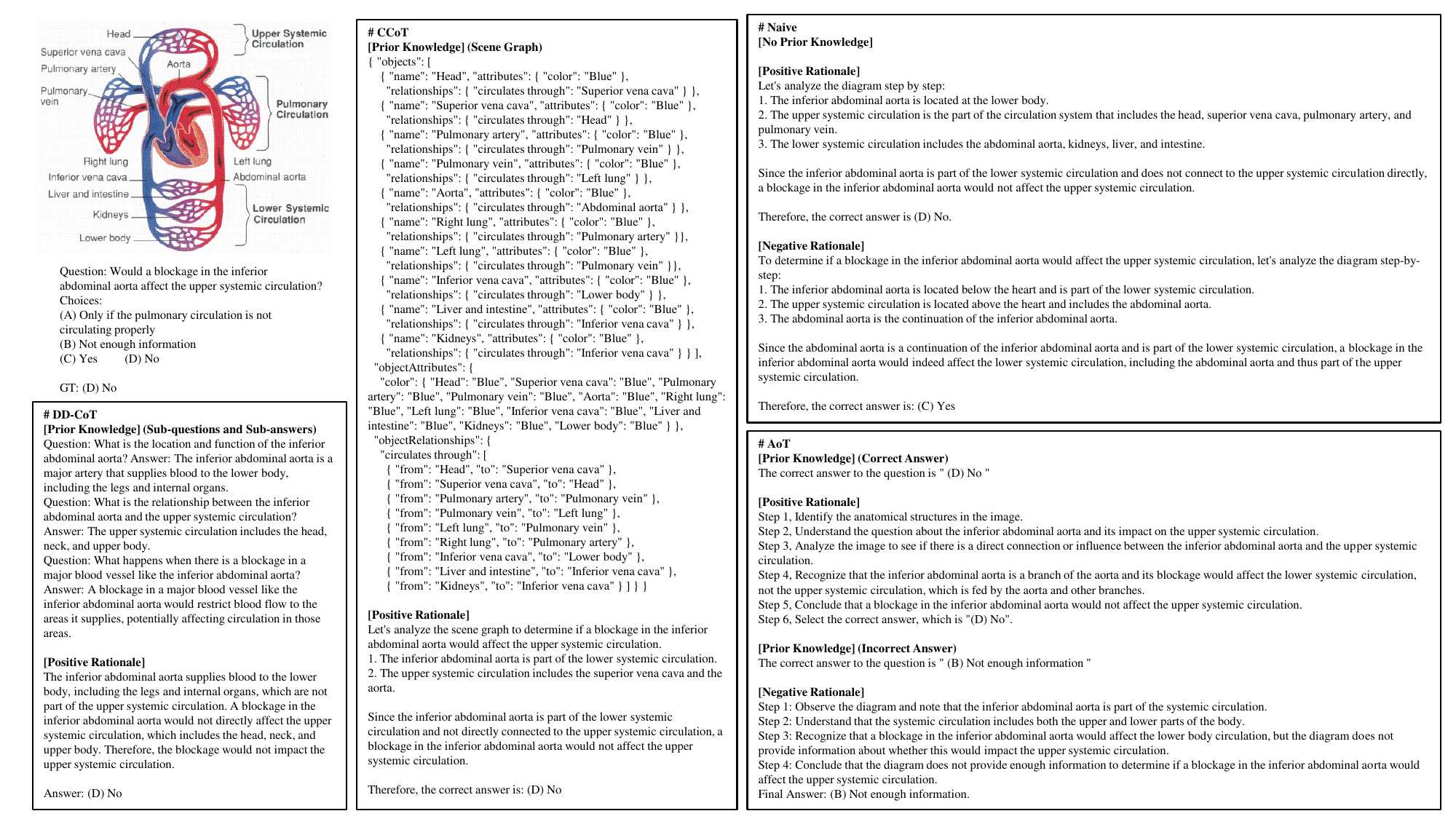}}
% \vspace{-1em}
\caption{Comparison of data generated by different prompts. We remove some symbols, such as $\$$ and $*$, for better reading.}
  \label{fig:figure_example4}
  % \vspace{-1em}
\end{figure*}

We present some examples generated by these prompts in Fig. \ref{fig:figure_example1}, \ref{fig:figure_example2}, \ref{fig:figure_example3}, and \ref{fig:figure_example4}, providing a detailed analysis.

\textbf{DD-CoT.} The rationales produced by DD-CoT rely heavily on prior knowledge. In Fig. \ref{fig:figure_example1}, the LLM effectively breaks down the question into necessary sub-questions, and the MLLM accurately extracts visual information, resulting in high-quality rationale. However, for more complex questions, such as those in Fig. \ref{fig:figure_example2} and \ref{fig:figure_example3}, it struggles to decompose the problems effectively, leading to missed critical information and reliance on guesswork for correct answers.

\textbf{CCoT.} CCoT can effectively extract scene graphs from images, typically providing key information such as the radius in Fig. \ref{fig:figure_example1}, and even the correct answer of 5/12 in Fig. \ref{fig:figure_example2}. However, they sometimes include redundant information, such as the coordinates of the four points of a square in Fig. \ref{fig:figure_example1}, or color attributes in Fig. \ref{fig:figure_example4}. Additionally, CCoT occasionally overlooks information already present in the scene graph during the reasoning process. For example, in Fig. \ref{fig:figure_example1}, CCoT ignores the radius already provided in the scene graph and extracts it again from the image. Optimizing the generation and utilization of scene graphs should better leverage the strengths of CCoT.

\textbf{Naive and AoT.} The Naive prompt derives answers directly without prior knowledge, while AoT utilizes correct and incorrect answers as prior knowledge. Although both can generate reasoning preference pairs, there are two notable differences in the data they produce:

(1) AoT typically generates more concise positive rationales. For instance, in Fig. \ref{fig:figure_example1}, Naive reasoning transitions from diameter to radius and then uses \(C = 2\pi r\) to calculate circumference, while AoT uses the more concise \(C = \pi d\), skipping the radius calculation step. The reason is that AoT knows the correct answer in advance, allowing it to accurately find the shortest solution path.

(2) AoT tends to produce negative reasoning with more errors. For example, in Fig. \ref{fig:figure_example2}, AoT makes mistakes in both the numerator and denominator, whereas Naive only miscalculates the numerator. More errors in negative examples are beneficial for DPO, as DPO works to decrease the likelihood of the model producing negative examples. As errors accumulate, the model becomes less likely to generate severe mistakes or hallucinations, ultimately enhancing its robustness.

These characteristics of AoT data also explain why the SMART model can produce more concise and accurate responses than the base model, as shown in Fig. 4 of the main text. This further confirms the effectiveness of using answers as hits.

\section{More Experiments} \label{sec:C}

\subsection{Performance Trends of Four Models} 
\begin{figure*}[]
    \centering
    
    \begin{subfigure}[t]{0.9\textwidth}
        \centering
        \includegraphics[width=\linewidth]{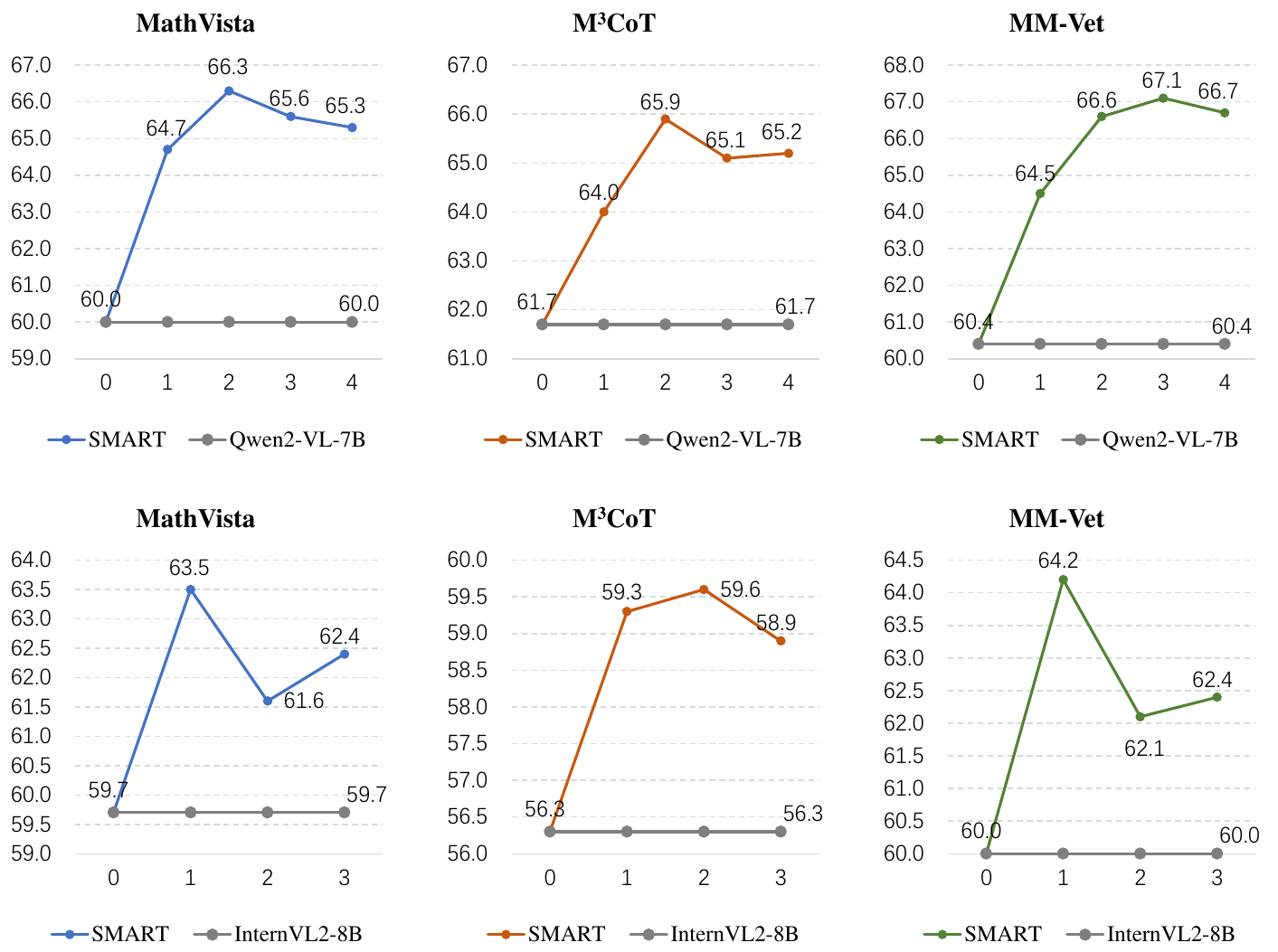} % 替换为您的图片1路径
        % \caption{Caption for Image 1} % 图片1的标题
        \label{fig:image1}
    \end{subfigure}
    
    % \vspace{0.5cm} % 图片之间的垂直间距，可以根据需要调整

    \begin{subfigure}[t]{0.9\textwidth}
        \centering
        \includegraphics[width=\linewidth]{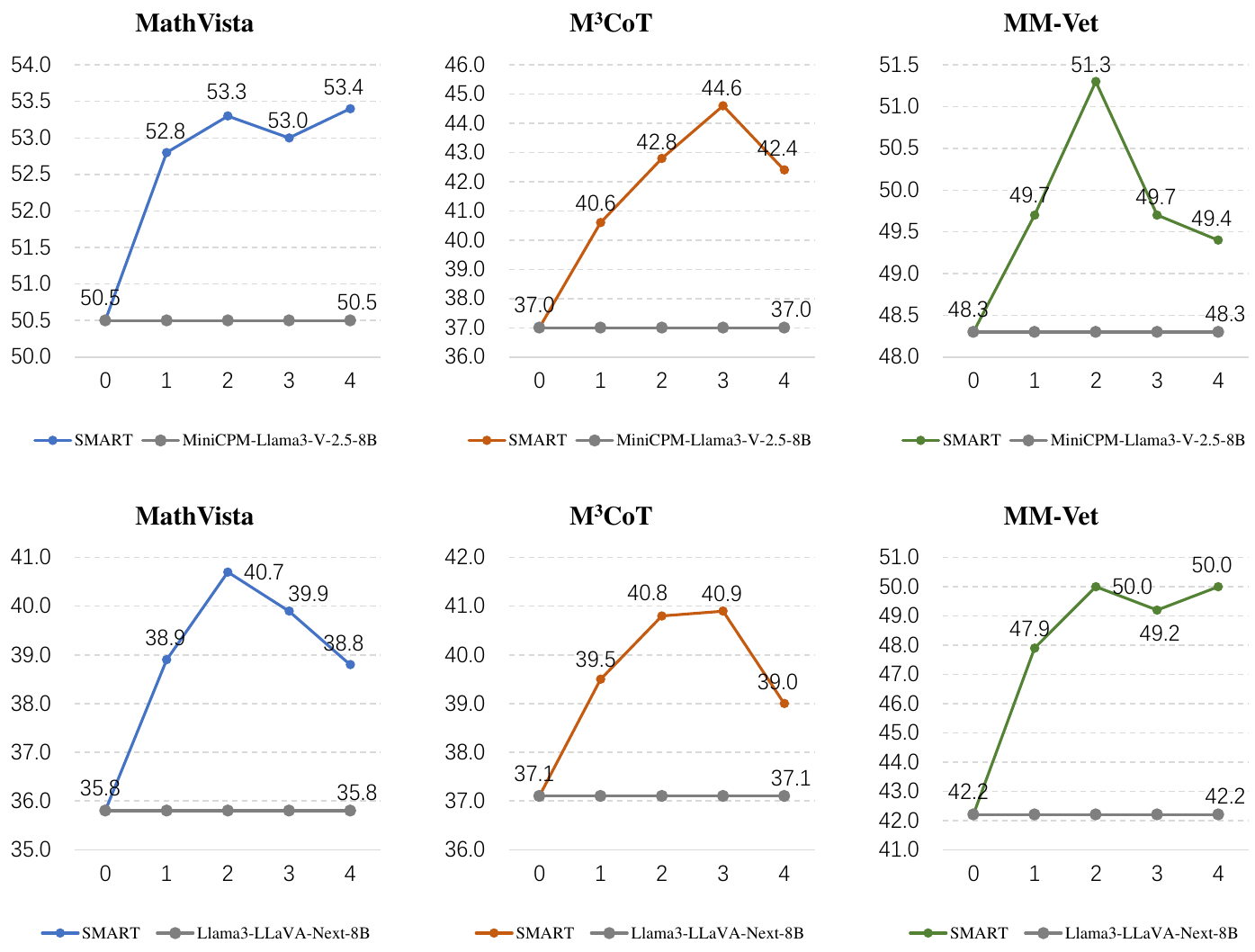} % 替换为您的图片2路径
        % \caption{Caption for Image 2} % 图片2的标题
        \label{fig:image2}
    \end{subfigure}
    
    \caption{Performance curve of the MLLMs across iterative preference alignment within the SMART framework. The model's performance quickly improves with more iterations but reaches saturation when it fully exploits its potential, consistent with the observations in \cite{pang2024iterative,tan2024beyond}.} % 整体标题
    \label{fig:two_images}
\end{figure*}

Fig. \ref{fig:two_images} illustrates the performance trajectories of four models, Qwen2-VL-7B \cite{Qwen2VL}, InternVL2-8B \cite{chen2024far}, MiniCPM-Llama3-V-2.5-8B \cite{yao2024minicpm}, and Llama3-LLaVA-Next-8B \cite{liu2024llavanext}, across multiple iterations of preference alignment within the SMART framework.
It shows that the performance increases with the number of iterations, validating the iterative ``generate-train" strategy. Notably, InternVL2-8B reaches performance saturation after just one iteration, while the other models benefit from up to two iterations, likely due to architectural or pre-training differences. Nevertheless, SMART proves effective across all four models, demonstrating robust generalizability.

\subsection{Further Comparisons with Advanced Datasets}
\begin{figure*}[]
\vspace{-1em}
\centerline{\includegraphics[width=0.975\linewidth]{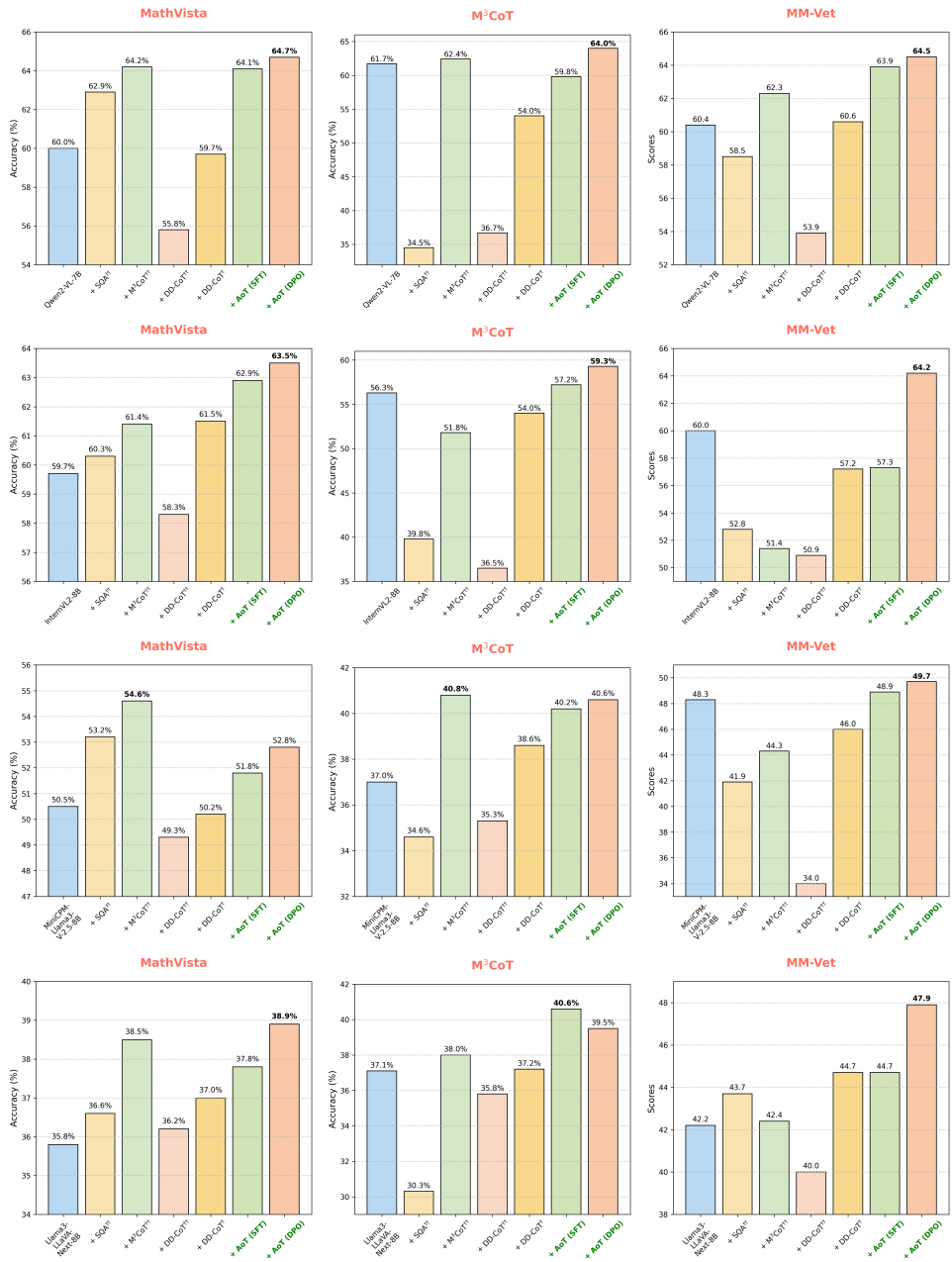}}

\caption{Performance comparison of various models fine-tuned on different reasoning datasets. The results highlight the high quality of AoT data.}
  \label{fig:figure_more_datasets}
\end{figure*}

In Table 3, we compare the fine-tuning results of Qwen2-VL-7B across various reasoning datasets, including the manually annotated SQA \cite{lu2022learn} and M$^3$CoT \cite{chen2024m} datasets, as well as the automatically annotated DD-CoT \cite{zheng2023ddcot} and our AoT data. To further validate our findings, we conduct the same experiments on three additional models: InternVL2-8B \cite{chen2024far}, MiniCPM-Llama3-V-2.5-8B \cite{yao2024minicpm}, and Llama3-LLaVA-Next-8B \cite{liu2024llavanext}. The results shown in Fig. \ref{fig:figure_more_datasets} reinforce our conclusions drawn in the main text: (1) AoT data represents the highest quality among currently available automatically generated datasets; (2) AoT is capable of generating negative examples, a feature overlooked by previous methods. By integrating negative examples and employing DPO function, AoT outperforms other methods in most scenarios, including those utilizing manually annotated datasets.

\subsection{Answer Hints Matter}
\begin{figure*}[]
% \vspace{-0.9em}
\centerline{\includegraphics[width=1.0\linewidth]{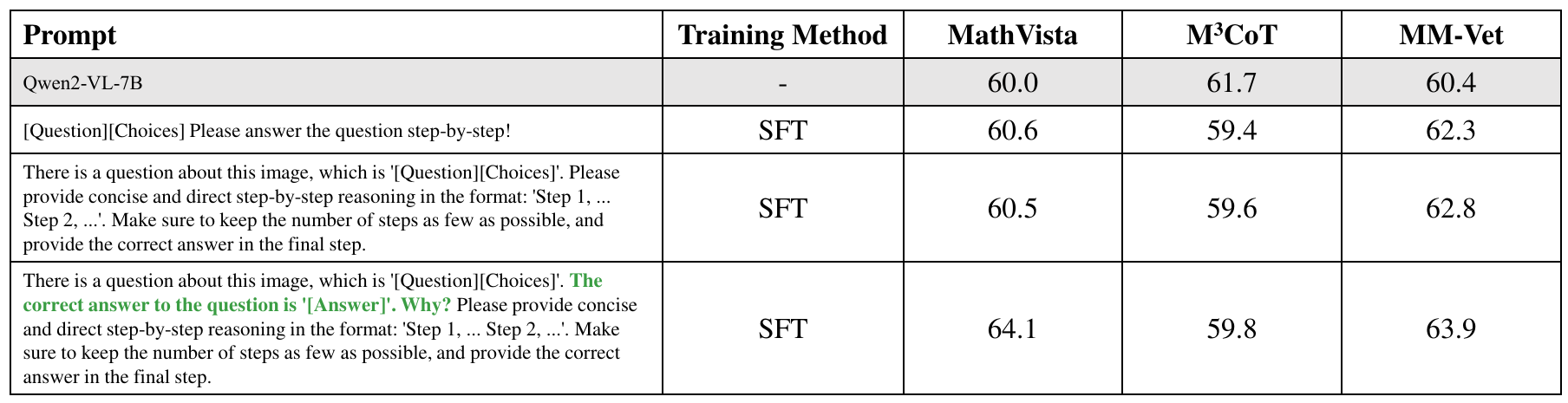}}
% \vspace{-1em}
\caption{Comparison of the performance of different prompts. Incorporating answers into the instructions is quite beneficial.}
  \label{fig:figure_vs_concise}
  % \vspace{-1em}
\end{figure*}

Multiple images suggest that the data generated using the AoT prompt is straightforward and often leads to solutions with fewer steps. To explore the role of the answer prior knowledge in this process, we remove the ``answer" hints from the AoT instructions while keeping the rest of the text unchanged. We then create 6K positive rationales to train the models, comparing these with models trained on the positive rationales generated by AoT. The results, shown in Fig. \ref{fig:figure_vs_concise}, demonstrate that the models trained with answer-guided data significantly outperform the others. This outcome supports the conclusions in Table 2 of the main text, highlighting the advantages of using answers as prior knowledge to produce high-quality reasoning data.

\subsection{More Evaluation Visualizations}
\begin{figure*}[]
% \vspace{-0.9em}
\centerline{\includegraphics[width=1.0\linewidth]{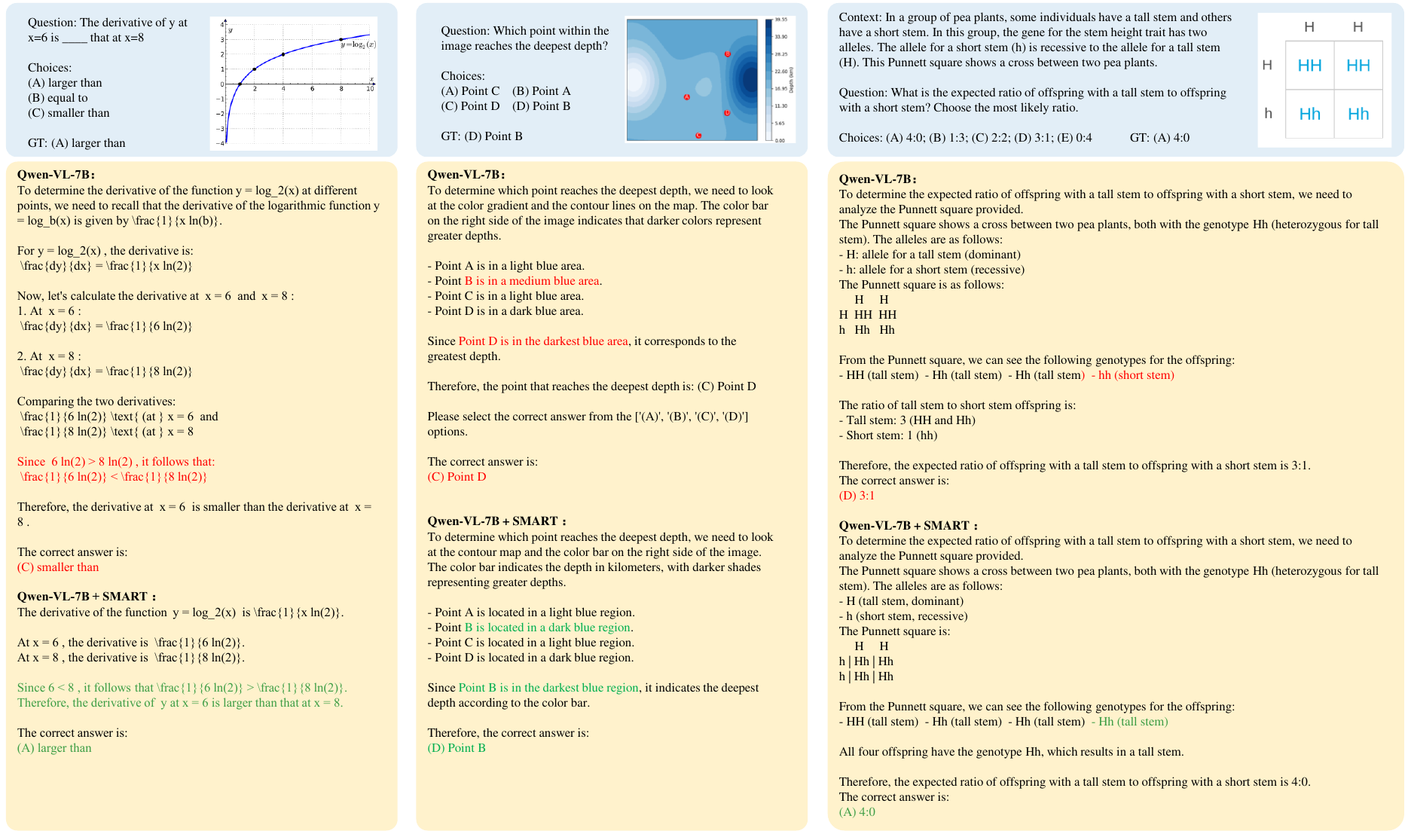}}
% \vspace{-1em}
\caption{Additional evaluation results from the Qwen2-VL-7B and SMART models, showcasing SMART's superior reasoning capabilities. To improve readability, we have removed certain symbols, such as $\$$ and $*$.}
  \label{fig:figure_more_visuals}
  % \vspace{-1em}
\end{figure*}

In this section, we present additional test results for the Qwen-VL-7B and SMART models in Fig. \ref{fig:figure_more_visuals}. Consistent with the conclusions drawn in Fig. 4, the SMART framework significantly enhances the reasoning capabilities of the Qwen-VL-7B model. For instance, in the mathematical problem involving derivatives (on the left), the Qwen-VL-7B model initially succeeded in its reasoning but made an error at a crucial step, arriving at an incorrect answer. In contrast, the SMART model reached the correct conclusion.

With the AoT improving the model's ability to extract visual information, the SMART model effectively utilized the color bar on the right side of the image during the subsequent depth comparison task (in the middle), accurately assessing the depth of each point to arrive at the correct answer. Similarly, in the biological question on the right, the model successfully extracted the information ``Hh (tall stem)" and answered the question correctly.

\end{appendices}


% \bibliography{sn-bibliography}% common bib file

\end{document}